# A two-stage denoising filter: the preprocessed Yaroslavsky filter [*]

Joseph Salmon[1], Rebecca Willett [1], and Ery Arias-Castro[2]

[1]Department of Electrical and Computer Engineering,
Duke University,
Durham, NC, USA.
[2]Department of Mathematics,
University of California,
San Diego, CA, USA.

September 3, 2012

## Abstract

This paper describes a simple image noise removal method which combines a preprocessing step with the Yaroslavsky filter for strong numerical, visual, and theoretical performance on a broad class of images. The framework developed is a two-stage approach. In the first stage the image is filtered with a classical denoising method (*e.g.,* wavelet or curvelet thresholding). In the second stage a modification of the Yaroslavsky filter is performed on the original noisy image, where the weights of the filters are governed by pixel similarities in the denoised image from the first stage. Similar prefiltering ideas have proved effective previously in the literature, and this paper provides theoretical guarantees and important insight into why prefiltering can be effective. Empirically, this simple approach achieves very good performance for cartoon images, and can be computed much more quickly than current patch-based denoising algorithms.

## 1 Introduction

This paper provides new insight into the performance of prefiltering steps used in many modern image denoising methods. Our analysis is inspired by recent results [1] characterizing the theoretical performance of neighborhood filters such as Yaroslavsky's filter (YF) [17] and non-local means

[*]The authors started working on the paper at the Institute for Mathematics and its Applications and gratefully acknowledge support from DARPA grant no. FA8650-11-1-7150, AFOSR award no. FA9550-10-1-0390, NGA award no. HM1582-10-1-0002, and NSF award no. CCF-06-43947.

(NLM) [2]. The approach in this paper consists of (1) applying a simple preprocessing step to the noisy image, and (2) using the preprocessed results to improve the weights used in Yaroslavsky's filter. The preprocessing step could correspond to a number of alternatives, such as linear filtering (LF) or wavelet [7] or curvelet [4] thresholding. Prefiltering in general been considered *empirically* in the development of non-local means with averages (as described in Section 3) [10], locally adaptive regression kernels [15], BM3D [6], and the gradient based anisotropic NLM in Maleki *et al.*[11], among others. However, little was understood on a theoretical level about the role and importance of prefiltering.

We refer to our two-stage method as the *preprocessed Yaroslavsky (PY) filter*. The main contributions in this paper are three-fold:

- Theoretical: we bound the decay of the MSE of the estimate as a function of the image smoothness and the number of pixels for "cartoon" images.
- Computational: we propose fast and simple algorithms for image denoising with few tuning parameters.
- Practical: the method introduced provides better performance on cartoon images (both numerically and visually) than any single method applied independently.

The remainder of the paper is organized as follows. Section 2 presents a mathematical formulation of the problem. Neighborhood filters are defined in Section 3. Section 4 introduces the PY filter we propose and describes its theoretical properties. Experiments are described in Section 5, illustrating the performance of the PY filter in practice.

## 2 Problem formulation

We cast the problem of image denoising as a non-parametric regression problem in the presence of white noise. We observe noisy samples $\{y_i \in \mathbb{R} : i \in I_n^d\}$ (with $I_n := \{1, \ldots, n\}$) of the target function $f : [0,1]^d \to [0,1]$ at the design points $\{x_i \in \mathbb{R}^d : i \in I_n^d\}$ corrupted by a zero mean additive white Gaussian noise with known variance $\sigma^2 > 0$, $\{\varepsilon_i \in \mathbb{R} : i \in I_n^d\}$, as follows

$$y_i = f(x_i) + \varepsilon_i, \quad i \in I_n^d. \tag{1}$$

(As noted in [1], many of our results hold for more general noise models as well.) We focus on a standard model in image processing where the sample points are on the square lattice, specifically, $x_i = ((i_1 - 1/2)/n, \ldots, (i_d - 1/2)/n)$ when $i = (i_1, \ldots, i_d)$. Leaving $n$ implicit, define vectors $\mathbf{y} = (y_i : i \in I_n^d)$, $\mathbf{f} = (f_i : i \in I_n^d)$ with $f_i := f(x_i)$ and $\boldsymbol{\varepsilon} = (\varepsilon_i : i \in I_n^d)$. The vector model can thus be written

$$\mathbf{y} = \mathbf{f} + \boldsymbol{\varepsilon}. \tag{2}$$

We assume that $f$ is a "cartoon image" – a piecewise smooth image with discontinuities along smooth hypersurfaces. For simplicity, we consider that $f$ is made of two pieces, with each piece being $\alpha$-Hölder smooth (as defined in [9]) and $\alpha \geq 1$ (*cf.* [1] the precise definition of $\mathcal{H}_d(\alpha, C_0)$, the set of $\alpha$-Hölder smooth functions with constant $C_0$).

**Definition 1 (Cartoon function class)** *For $\alpha, C_0 > 0$, let $\mathcal{F} = \mathcal{F}(\alpha, C_0)$ denote the set of functions of the form*

$$f(x) = \mathbb{1}_{\{x \in \Omega\}} f_\Omega(x) + \mathbb{1}_{\{x \in \Omega^c\}} f_{\Omega^c}(x), \tag{3}$$

where $f_\Omega, f_{\Omega^c} \in \mathcal{H}_d(\alpha, C_0)$, with jump (or discontinuity gap)

$$\mu(f) := \inf_{x \in \partial\Omega} |f_\Omega(x) - f_{\Omega^c}(x)| \geq 1/C_0, \tag{4}$$

and $\Omega \subset (0,1)^d$ is a bi-Lipschitz image of the (Euclidean) unit ball $B(0,1)$, specifically, $\Omega = \phi(B(0,1))$, where $\phi : \mathbb{R}^d \to \mathbb{R}^d$ is injective with $\phi$ and $\phi^{-1}$ both Lipschitz with constant $C_0$ (i.e., $C_0$-Lipschitz) with respect to the supnorm. We refer to $f_\Omega$ as the foreground and to $f_{\Omega^c}$ as the background. Moreover $\partial\Omega$ represents the (topological) boundary of $\Omega$.

The condition (4) is a lower bound on the minimum "jump" along the discontinuity $\partial\Omega$. We require that $\phi$ is bi-Lipschitz to ensure that the set $\Omega$ is sufficiently smooth and does not have a serious bottleneck, which could potentially mislead the methods discussed here.

Our goal is to estimate the vector $\mathbf{f}$ and we measure the performance of an estimator $\widehat{\mathbf{f}}$ in terms of mean square error (MSE):

$$\text{MSE}_f(\widehat{\mathbf{f}}) = \frac{\mathbb{E}\|\widehat{\mathbf{f}} - \mathbf{f}\|_2^2}{n^d} = \frac{1}{n^d} \sum_{i \in I_n^d} \mathbb{E}(\widehat{f}_i - f_i)^2,$$

where the expectation $\mathbb{E}$ is with respect to the probability measure associated with the noise. In particular, we are interested in understanding the worse-case MSE performance of potential denoising methods, as measured by

$$\mathcal{R}_n := \sup_{f \in \mathcal{F}} \text{MSE}_f(\widehat{\mathbf{f}}).$$

## 3 Neighborhood filters

We consider neighborhood filters of the form

$$\widehat{f}_i = \frac{\sum_{j \in I_n^d} \omega_{i,j}\, y_j}{\sum_{k \in I_n^d} \omega_{i,k}}. \tag{5}$$

where the weights $\omega_{i,j}$ (may) depend on the observation $\mathbf{y}$. This general class of filters has recently been thoroughly studied in the literature (*cf.* [14] and [13])

The methods we study differ only in the weights $\omega_{i,j}$ used. For $\alpha > 1$ we use a particular version of (5) which incorporates local polynomial regression (LPR), as detailed in [1]. This allows us to adapt to higher orders of smoothness without altering the kernels used below.

**Linear filtering (LF):** In this context the weights are controlled by spatial proximity only. Using a kernel $K$ and a bandwidth $h > 0$, the weights can be written

$$\omega_{i,j} = K_h(x_i, x_j), \tag{6}$$

where $K_h(x_i, x_j) = K(\frac{x_i}{h}, \frac{x_j}{h})$ for any sample points $x_i$ and $x_j$.

**Weight oracle (WO):** We now introduce an oracle estimator, the *weight oracle*, which can choose the weights based on the true image $f$:

$$w_{i,j} := K_h(x_i, x_j)\mathbb{1}\{|f_i - f_j| < h_y\}. \tag{7}$$

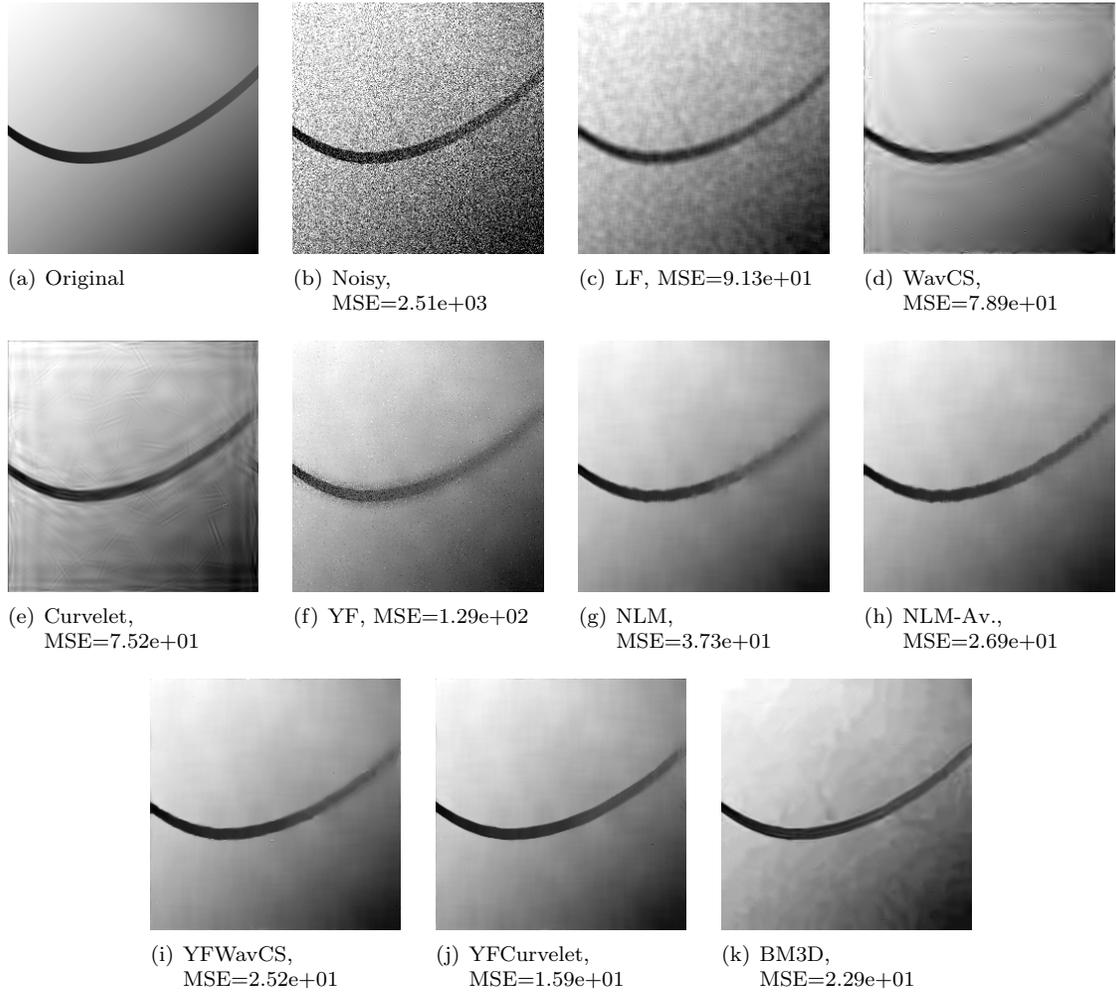

Figure 1: Toy cartoon image (Swoosh) corrupted with Gaussian noise with $\sigma = 50$.

As before, $K$ and $h$ control the spatial proximity; the photometric bandwidth $h_y$ controls the photometric proximity. This oracle is closely related to the *membership oracle* introduced in our previous work [1] and shares the same performance characteristics. In particular, we have the following:

**Theorem 1 (Weight oracle)** *Let $\widehat{\mathbf{f}}_h^{\mathrm{WO}}$ denote a neighborhood filter* (5) *using LPR with weights as in* (7). *Then*
$$\inf_h \mathcal{R}_n(\widehat{\mathbf{f}}_h^{\mathrm{WO}}, \mathcal{F}) \asymp \mathcal{R}^{\mathrm{WO}} := (\sigma^2/n^d)^{2\alpha/(d+2\alpha)},$$
*and the optimal choice of bandwidths are $h \asymp h^{\mathrm{WO}} := (\sigma^2/n^d)^{1/(d+2\alpha)}$ and $h_y \asymp 1$.*

*Proof:* The proof follows the proof of Theorem 4.4 in [1]. ∎

**Yaroslavsky's filter (YF):** For the YF [17], the similarity between two pixels is based on their spatial distance and on the relative proximity of image intensity at these pixels:

$$\omega_{i,j} = K_h(x_j, x_j) \, \mathbb{1}\{|y_i - y_j| < h_y\}. \tag{8}$$

We showed in [1] that when the noise is sufficiently small (*i.e.*, $\sigma^2 = O(1/\sqrt{\log n})$), then $|y_i - y_j|$ is a close approximation to $|f_i - f_j|$, and the YF performs nearly as well as the weight oracle described above. However, when the noise is strong this approach is fragile.

**Non-Local Means (NLM):** The fragility of the YF led to the development of *nonlocal means (NLM)*, in which one estimates the photometric distance between pixels using patches of noisy pixels [2]. For $h_\mathsf{P} > 0$ let $\mathbf{y}_{\mathsf{P}_i} = (y_j : \|x_j - x_i\| \leq h_\mathsf{P})$ be the vector of pixel values over the patch centered at $x_i$. The weights used in NLM are:

$$\omega_{i,j} = K_h(x_i, x_j) \, \mathbb{1}\{\|\mathbf{y}_{\mathsf{P}_i} - \mathbf{y}_{\mathsf{P}_j}\| < h_y\}. \tag{9}$$

**Non-Local Means Average (NLM-Av.):** A drawback of NLM is the large computation time associated with computing the distances between patches in (9). Empirical evidence [10] and theoretical results [16, 1]) show that a fast approximation to NLM is also effective on cartoon images. This approximation amounts to computing the average of pixels within each patch, and using the differences of the averages to estimate photometric distances. We refer to this method as NLM-Av. :

$$\omega_{i,j} = K_h(x_i, x_j) \, \mathbb{1}\{|\overline{y}_{\mathsf{P}_i} - \overline{y}_{\mathsf{P}_j}| < h_y\}, \tag{10}$$

where $\overline{y}_{\mathsf{P}_i}$ is the pixel average on patch $i$.

## 4 The Preprocessed Yaroslavsky (PY) Filter

### 4.1 Proposed algorithm

As seen above, when the true image intensity is used to compute the weights in a neighborhood filter, we achieve very strong performance guarantees. This suggests the following two-stage approach:

1. Compute an initial estimate of $\mathbf{f}$, denoted $\widetilde{\mathbf{f}} := \text{denoise}(\mathbf{y})$.

2. Use $\widetilde{\mathbf{f}}$ to compute the weights in a Yaroslavsky-type filter

$$w_{i,j}^{\text{PY}} := K_h(x_i, x_j) \mathbb{1}\{|\widetilde{f}_i - \widetilde{f}_j| < h_y^{\text{PY}}\}. \tag{11}$$

We call our approach a *Preprocessed Yaroslavsky (PY)* filter. NLM-Av. is an example of the two-stage approach above where $\widetilde{\mathbf{f}}$ corresponds passing $\mathbf{y}$ through a linear boxcar filter with sidelength $h_\mathsf{P}$. In this case, $\widetilde{\mathbf{f}}$ is a poor estimate of $\mathbf{f}$, but enough of an improvement over the raw data $\mathbf{y}$ that it can be used to compute effective weights. *The key idea is that if $\widetilde{\mathbf{f}}$ is a good estimate of $\mathbf{f}$, then computing weights using $\widetilde{\mathbf{f}}$ will closely approximate the WO above. We show that this is true with theoretical analysis and simulations.*

NLM-Av. is a good estimator only in the smooth regions. Close to a boundary it is well known that wavelet and curvelet [7, 4, 3] thresholding will provide better $\tilde{\mathbf{f}}$ in (11). Indeed, we find that using wavelet or curvelet thresholding to estimate $\widetilde{\mathbf{f}}$ and then using (11) results in a strong estimator with many fewer artifacts than we see with $\widetilde{\mathbf{f}}$ alone. For cartoon images we often *outperform NLM with a small fraction of its computation time*, and sometimes outperform BM3D [6], a state-of-the-art image denoising algorithm currently without theoretical support. However, for textures and natural images the more sophisticated BM3D outperforms the proposed method.

## 4.2 Performance bounds

We are able to characterize the global performance of our proposed PY using deviation bounds on the preprocessed estimator $\tilde{\mathbf{f}}$.

**Theorem 2** *Suppose an estimator $\widetilde{\mathbf{f}}$ satisfies for any $f \in \mathcal{F}$ the following deviation bound, with probability at least $1 - \delta$:*

$$|\widetilde{f}_i - f_i|^2 \leq M, \qquad \forall i \in I_n^d \text{ such that } B(x_i, \widetilde{h}) \cap \partial\Omega = \emptyset. \qquad (12)$$

*Then, if $M = o(1)$, for $\widehat{\mathbf{f}}_h^{\text{PY}}$ defined with weights as in (11), one has*

$$\inf_h \mathcal{R}_n(\widehat{\mathbf{f}}_h^{\text{PY}}, \mathcal{F}) \asymp \widetilde{h} + \delta + (\sigma^2/n^d)^{2\alpha/(d+2\alpha)},$$

*and the optimal choice of bandwidths are $h \asymp h^{\text{WO}}$ and $h_y \asymp 1$.*

**Remark 1** The deviation bounds in (12) means that for a distance greater than $\widetilde{h}$ away from the boundary, the error is bounded by $M$. Near the boundary the error is bounded by 1 (since we consider bounded $f$ and $\widetilde{f}$).

**Remark 2** Note that in order to achieve the optimal rate, $\widetilde{h}$ must decay no more slowly than $(\sigma^2/n^d)^{2\alpha/(d+2\alpha)}$; thus, for this method to work effectively, both $M$ and $\widetilde{h}$ must simultaneously decay sufficiently quickly. In the case of $\widetilde{\mathbf{f}}$ corresponding to a linear smoothing filter, choosing $\widetilde{h}$ too large will hurt the above MSE, but at the same time if it is chosen to be too small then the corresponding $M$ will be large and we will be unable to effectively mimic the weight oracle.

*Proof:* Let $E$ denote the event that (12) holds; note $\mathbb{P}(E) \geq 1 - \delta$ and

$$\text{MSE}_f(\widehat{\mathbf{f}}_i) = \frac{\mathbb{E}\left(\|\mathbf{f} - \widehat{\mathbf{f}}\|_2^2 | E\right)}{n^d}\mathbb{P}(E) + \frac{\mathbb{E}\left(\|\mathbf{f} - \widehat{\mathbf{f}}\|_2^2 | E^c\right)}{n^d}\mathbb{P}(E^c) \leq \frac{\mathbb{E}\left(\|\mathbf{f} - \widehat{\mathbf{f}}\|_2^2 | E\right)}{n^d} + \delta.$$

For $B(x_i, \widetilde{h}) \cap \partial\Omega \neq \emptyset$ (*i.e.*, for points near a boundary), we have $|f_i - \widetilde{f}_i| \leq 1$; note that there are $O(n^d\widetilde{h})$ such points. Now consider $i$ such that $B(x_i, \widetilde{h}) \cap \partial\Omega = \emptyset$. Conditioning on $E$ and applying the triangle inequality:

$$|f_i - f_j| - 2\sqrt{M} \leq |\widetilde{f}_i - \widetilde{f}_j| \leq |f_i - f_j| + 2\sqrt{M}$$

Since $M = o(1)$, when $E$ holds the photometric kernel in the preprocessed Yaroslavsky filter is able to exactly mimic the *weight oracle*. Summing over boundary and non-boundary pixels and dividing by $n^d$, the overall MSE is bounded by $\widetilde{h} + (\sigma^2/n^d)^{2\alpha/(2\alpha+d)} + \delta$. ∎

**Example 1 (Linear filter [1])** *If $\widetilde{\mathbf{f}}$ corresponds to convolving the noisy image $\mathbf{y}$ with a smooth kernel with bandwidth $\widetilde{h}$, then $\widetilde{\mathbf{f}}$ satisfies the condition (12) for $M = C_0\tilde{h}^{2\alpha} + C\sigma^2 \log(n^d)/(n\tilde{h})^d$ and $\delta = (n\tilde{h})^{(d(1-C))}$.*

| Size | LF | WavCS | Curv. | YF | YFWavCS | YFCurv. | NLM-Av | NLM | BM3D |
|---|---|---|---|---|---|---|---|---|---|
| $256^2$ | 0.03 s | 0.08 s | 0.33 s | 0.16 s | 0.26 s | 0.48 s | 0.15 s | 14.75 s | 1.18 s |
| $512^2$ | 0.08 s | 0.53 s | 1.13 s | 0.72 s | 1.28 s | 1.75 s | 0.63 s | 60.00 s | 4.99 s |
| $1024^2$ | 0.18 s | 2.97 s | 3.90 s | 2.89 s | 5.94 s | 6.47 s | 2.48 s | 241.9 s | 21.4 s |

Figure 2: Computing times for Matlab mex/C implementations (except the Matlab script for the curvelet transform) on an Intel Core i7 CPU 2.67GHz.

We are unaware of explicit deviation bounds for wavelet and curvelet thresholding methods that would hold in the cartoon model. However, work by Hong and Birget [8] (for Hölder functions in bounded noise, with $d = 1$) suggests that such bound are possible.

## 5 Experiments

Limitations and artifacts associated with common methods such as wavelet and curvelet thresholding appear when the noise level is strong; see Figs. 1c and 1d for illustrations. With wavelets, grainy outliers often appears in smooth regions, whereas with curvelets, artificial elongated stripes can be perceived throughout in the image. The YF performs almost optimally when the noise level is small; however, when the noise is strong, a lot of pixels are left unmodified. This leads to images with visible residual noise, *cf.* Fig. 1e, while our method can soften those visual degradations (*cf.* Fig. 1 h,i,j).

We perform comparisons on toy images and on natural images where the noise is Gaussian with known variance $\sigma^2$. We focus on the following methods:

- LF (Linear filtering)

- WavCS (Haar Wavelet with hard thresholding [7] and cycle spinning [5]),

- Curvelet (Curvelet with hard thresholding [3]),

- YF (Yaroslavsky Filter [17]),

- PYWavCS (Preprocessed Yaroslavsky filter with wavelet and cycle spinning),

- PYCurvelet (Preprocessed Yaroslavsky filter with curvelet),

- PYLF (Preprocessed Yaroslavsky filter with linear filter, equivalent to NLM Av.),

- NLM (Non-Local Means [2]),

- BM3D (a state-of-the-art method [6], with default parameters),

- WO (Weight Oracle).

The spatial kernel is the box kernel. When needed, the spatial bandwidth is $h^d = 21^2$ and patches have sidelength $h_\mathsf{P} = 7$. Performance is summarized in Table 3, with the following parameters:

- $\tau_{\mathrm{HT}}^{\mathrm{W}} = 3.5\sigma$ (WavCS, PYWavCS)
- $\tau_{\mathrm{HT}}^{\mathrm{C}} = 3\sigma$ (Curvelet, PYCurvelet)

|            | Blob   | Swoosh | Ridges | Cameraman |
|------------|--------|--------|--------|-----------|
|            |        | $\sigma = 5$ |   |           |
| LF         | 35.33  | 40.29  | 48.80  | 437.79    |
| WavCS      | 1.40   | 1.78   | 1.65   | 14.74     |
| Curvelet   | 5.12   | 4.88   | 1.58   | 28.96     |
| YF         | 1.27   | 2.10   | 16.95  | 14.57     |
| NLM-Av.    | 0.86   | 2.39   | 4.19   | 315.96    |
| YFWavCS    | **0.78** | 1.21 | 1.49   | 14.12     |
| YFCurvelet | 1.16   | 2.02   | 1.81   | 24.42     |
| NLM        | 0.96   | **1.14** | 2.11 | 13.40     |
| BM3D       | 1.22   | 1.20   | **0.88** | **9.95** |
| WO         | 1.61   | 2.15   | 36.36  | 32.59     |
|            |        | $\sigma = 20$ |   |           |
| LF         | 43.19  | 48.17  | 56.62  | 445.65    |
| WavCS      | 15.31  | 20.15  | 14.98  | 102.07    |
| Curvelet   | 19.20  | 34.32  | 10.66  | 148.92    |
| YF         | 17.00  | 22.00  | 189.05 | 120.11    |
| NLM-Av.    | 5.06   | 7.39   | 18.95  | 345.69    |
| YFWavCS    | 4.19   | 6.41   | 13.57  | 88.76     |
| YFCurvelet | **3.22** | **4.67** | 13.88 | 114.92 |
| NLM        | 4.03   | 4.74   | 25.98  | 91.72     |
| BM3D       | 5.72   | 7.04   | **8.94** | **59.36** |
| WO         | 2.66   | 3.19   | 38.00  | 34.24     |
|            |        | $\sigma = 50$ |   |           |
| LF         | 87.27  | 92.32  | 100.64 | 489.74    |
| WavCS      | 52.90  | 79.03  | 72.06  | 286.36    |
| Curvelet   | 51.73  | 75.46  | 50.91  | 290.34    |
| YF         | 106.16 | 126.71 | 547.43 | 501.54    |
| NLM-Av.    | 22.63  | 26.46  | 95.96  | 412.52    |
| YFWavCS    | 21.90  | 26.32  | 106.70 | 236.28    |
| YFCurvelet | **15.89** | **15.55** | 77.27 | 229.25 |
| NLM        | 28.70  | 35.46  | 209.34 | 266.40    |
| BM3D       | 18.52  | 23.71  | **36.30** | **158.35** |
| WO         | 8.47   | 9.00   | 47.40  | 43.42     |

Figure 3: MSE comparisons, with results averaged over 100 Gaussian noise replicas. Images are the same as in [1].

- $h_y = 0.2\sigma$ (PYLF, PYCurvelet)
- $h_y = \sqrt{10}\sigma$ (YF)
- $h_y^{\text{ORACLE}} = 30$ (WO)

We have also report computing times in Table 2 to illustrate the speed of the algorithms at stake. Fast transforms are available for wavelets [12] and curvelets [3]. The YF is faster than the NLM: the naive implementation of the YF has a computational complexity of $O(n^d h^d)$ while for NLM is $O(n^d h^d h_{\mathsf{P}}^d)$. For neighborhood filters such as YF or NLM, parallelization can exploit modern architectures. All the PY methods tested were two to eight times faster than BM3D.

The code used for these experiments is available on the authors' webpage: [http://josephsalmon.eu/code/index_codes.php?page=Neighborhood_filters](http://josephsalmon.eu/code/index_codes.php?page=Neighborhood_filters).

# 6  Conclusion

We have proposed and analyzed both theoretically and in practice the performance of a two-stage method based on preprocessing to determine the weights in a Yaroslavsky filter. This procedure behaves particularly well on cartoon images, reducing common artifacts produced either by the Yaroslavsky Filter or wavelet or curvelet thresholding. Moreover, it has the benefit of being reasonably fast with respect to patch-based methods (such as NLM [2] and BM3D [6]) and requires very few parameters to be tuned.